\documentclass[11pt,a4paper]{article}
\usepackage{times,latexsym}
\usepackage{url}
\usepackage[T1]{fontenc}

\usepackage[acceptedWithA]{tacl2021v1}

\usepackage{xspace,mfirstuc,tabulary}
 \usepackage{adjustbox}
 \usepackage{graphicx}
 \usepackage{subcaption}
\usepackage{makecell}

\newif\iftaclinstructions
\taclinstructionsfalse 
\iftaclinstructions
\renewcommand{\confidential}{}
\renewcommand{\anonsubtext}{(No author info supplied here, for consistency with
TACL-submission anonymization requirements)}
\newcommand{\instr}
\fi

\iftaclpubformat 

\else

\fi

\newcommand{\myset}[1]{\left\{#1\right\}}
\newcommand{\paren}[1]{\left(#1\right)}

\newcommand{\abs}[1]{\left|#1\right|}

\newcommand{\inner}[2]{\left\langle #1, #2 \right\rangle}

\newcommand{\diag}[1]{\mathrm{diag}\paren{#1}}

\newcommand{\enc}{\mathrm{enc}}
\newcommand{\score}{\mathrm{s}}
\newcommand{\pop}{\textbf{pop}}

\newcommand{\si}{\ensuremath{\sigma}}

\newcommand{\be}{\ensuremath{\beta}}
\newcommand{\R}{\ensuremath{\mathbb{R}}}

\newcommand{\ra}{\ensuremath{\rightarrow}}

\usepackage{mathtools}

\newcommand{\by}{\times}

\usepackage{amsmath,amssymb,amsthm,bbm}
\usepackage{algorithm,algpseudocode}
\DeclareMathOperator*{\argmax}{arg\,max}

\title{Improving Multitask Retrieval by Promoting Task Specialization}

\author{
      Wenzheng Zhang$^1$\thanks{\;\; Work done during an internship at Microsoft}\hspace{10mm}  Chenyan Xiong$^2$\hspace{10mm} Karl Stratos$^1$ \hspace{10mm} Arnold Overwijk$^2$ \\
  $^1$ Rutgers University, USA \hspace{5mm} $^2$ Microsoft, USA \\
  \texttt{\{wenzheng.zhang, karl.stratos\}@rutgers.edu} \\
  \texttt{\{chenyan.xiong, arnold.overwijk\}@microsoft.com}
}

\date{}

\begin{document}
\maketitle
\begin{abstract}
  In multitask retrieval, a single retriever is trained to retrieve relevant contexts for multiple tasks.
  Despite its practical appeal, naive multitask retrieval lags behind task-specific retrieval in which a separate retriever is trained for each task.
  We show that it is possible to train a multitask retriever that outperforms task-specific retrievers by promoting task specialization.
  The main ingredients are: (1) a better choice of pretrained model---one that is explicitly optimized for multitasking---along with compatible prompting,
  and (2) a novel adaptive learning method that encourages each parameter to specialize in a particular task.
  The resulting multitask retriever is highly performant on the KILT benchmark.
  Upon analysis, we find that the model indeed learns parameters that are more task-specialized compared to naive multitasking without prompting or adaptive learning.\footnote{Our code and model checkpoints are publicly available at \url{https://github.com/WenzhengZhang/TACO}.}
\end{abstract}

\section{Introduction}

A standard approach to knowledge-intensive language tasks such as question answering (QA), entity disambigution, and fact verification is retrieval-based.
Given an query, a retriever is used to efficiently search a large knowledge base (KB) to retrieve relevant ``contexts'', typically in the form of short paragraphs.
How these contexts are used is task-specific (e.g., entity disambiguation takes the title of the article in which the top retrieved context is found;
QA predicts an answer from the contexts by through a reader model).
In this paper, we focus on the retrieval step.

In particular, we focus on multitask retrieval. In this setting, there are $K > 1$ downstream tasks that benefit from retrieval from a shared KB.
A single retriever is then tasked with performing retrieval for $K$ tasks.
Multitask retrieval contrasts with task-specific retrieval in which a separate retriever is trained for each task,
and has compelling advantages such as model simplicity (i.e., we can use the same model for all tasks rather than having to design potentially different models for different tasks)
and memory efficiency at test time ($K$ times smaller).

Despite the practical appeal, the performance of multitask retrieval has been underwhelming, severely limiting its real-world applicability.
Specifically, previous work by \citet{maillard2021multi} trains DPR \citep{karpukhin-etal-2020-dense} on the union of all training datasets in the KILT benchmark \citep{petroni2021kilt},
but the model is outperformed by task-specific retrievers in 5 out of 8 tasks (page-level $R$-precision, validation split).
In our experiments, we find that it is in fact outperformed in all tasks (often by substantial margins) when a stronger task-specific baseline is used.
This result is surprising as well as disappointing given the usual benefits of multitask learning (e.g., data efficiency, reduced overfitting) when properly done.

We debunk the previous negative result by presenting a multitask retriever that outperforms task-specific retrievers.
The main theme of our work is that it is beneficial to explicitly promote task specialization.
A first important source of improvement is a better choice of pretrained model, one that is explicitly optimized for multitasking.
Specifically, instead of the standard retrieval encoder BERT \citep{devlin2019bert}, we use T5 \citep{raffel2019t5} which includes multitasking in its pretraining stage.
Importantly, we use the same prompting as in pretraining (i.e., task indicator) to reduce the gap between pretraining and finetuning for multitask retrieval.
A second source of improvement is a novel adaptive learning method in which we adatively upweight the task gradients by the parameter's sensitivity to these tasks to encourage task specialization.

The resulting multitask retriever is highly performant on the KILT benchmark.
We achieve 73.74\%  average page-level R-precision on KILT validation data and  72.84\% average page-level R-precision on KILT test data.
Upon analysis, we find that the model indeed learns parameters that are more task-specialized compared to naive multitasking without prompting or adaptive learning.

\section{Related Work}

\citet{maillard2021multi} propose multitask retrieval largely as an extension of DPR.
Their best model is a BERT-based dual encoder trained on the union of 8 retrieval tasks.
While it performs comparably with task-specific DPRs on some tasks, it generally lags behind.
In this work, we use stronger task-specific retrievers based on T5 and ANCE \citep{xiong2020approximate}
all of which substantially outperform their multitask retriever.
We argue that this negative result undermines the case for multitask retrieval and that it is crucial to demonstrate competitive performance.
Our main contribution is producing this demonstration.

We emphasize that achieving competitive multitask retrieval in practice is a highly difficult empirical problem.
One might think that it is simply an application of multitask learning, which has no shortage of sophisticated techniques.
These techniques typically modify the gradients during training, such as gradient surgery \citep{yu2020gradient}, gradient vaccine \citep{wang2020gradient},
common gradient descent \citep{piratla2021focus}, and GradNorm \citep{chen2018gradnorm}.
We experiment with these techniques and find that they do not help, thus motivated to develop one that works.

Our technical contribution is a new method for multitask learning based on the notion of task sensitivity.
Given a loss function $J(\theta)$, the sensitivity of the $i$-th parameter to the loss at $\theta$ is
defined as the absolute change in the loss when $\theta_i$ is set to zero,
which can be approximated by a first-order Taylor approximation as
\begin{align*}
  \abs{J(\theta) - J(\theta_{-i})} \approx \abs{\frac{\partial J(\theta)}{\partial \theta_i} \times \theta_i}
\end{align*}
where $\theta_{-i}$ is equal to $\theta$ except that its $i$-th element is zero.
This quantity has been used in the context of model pruning---as a way of identifying weakly sensitive weights \citep{molchanov2016pruning,molchanov2019importance,michel2019sixteen,liang2021super}
and updating them more aggresively \citep{liang2022no}.
In contrast, we use the quantity to identify weights that are strongly sensitive to a particular task and increase their sensitivity even further, intuitively to achieve per-parameter task specialization.
To our knowledge, we are the first to use parameter sensitivity for multitasking learning.

We briefly differentiate our work from other recent works on multitask retrieval.
\citet{chen2022corpusbrain} present CorpusBrain, an autoregressive multitask retriever trained in largely the same style as GENRE \citep{cao2021autoregressive} with excellent performance.
Autoregressive retrieval has different pros and cons compared to dense retrieval which is our setting;
it can be more memory and runtime efficient, but it does not ``read'' the description of the target and thus not suited for retrieval tasks that require involved reasoning over query-target pairs (e.g., zero-shot entity retrieval \citep{logeswaran2019zero}).
Thus we consider the contribution of CorpusBrain to be at least partially orthogonal to ours. Nevertheless, we show that our model outperforms CorpusBrain in a similar training setting in experiments.
\citet{asai2022task} propose instruction-based retrieval in which the retriever is given an intent as well as a query to find the intended target.
While this is a form of multitask retrieval, the problem formulation is different and it is evaluated on its own dataset benchmark.

\section{Method}

We build on the well-established framework of dual encoder \citep[\textit{inter alia}]{bromley1993signature,huang2013learning,gillick-etal-2019-learning,karpukhin-etal-2020-dense}.
Let $\mathcal{X}$ denote the set of all queries and $\mathcal{Y}$ the set of all targets (i.e., KB).
First, we assume mappings $\mathrm{text}_X: \mathcal{X} \ra \mathcal{V}^+$ and $\mathrm{text}_Y: \mathcal{Y} \ra \mathcal{V}^+$ where $\mathcal{V}$ denotes the vocabulary
to ``verbalize'' queries and targets.
Second, we assume encoders $\enc^\theta_X, \enc^\theta_Y: \mathcal{V}^+ \ra \R^d$ with parameters $\theta$
defining the relevance score function $\score_\theta(x,y) = \inner{\enc^\theta_X(\mathrm{text}_X(x))}{\enc^\theta_Y(\mathrm{text}_Y(y))}$.
Third, assuming iid samples $(x_1, y_1) \ldots (x_N, y_N) \sim \pop$, we learn the parameters by noise contrastive estimation (NCE):
\begin{align*}
  \min_\theta\; -\frac{1}{N} \sum_{i=1}^N \log \frac{ \exp(\score_\theta(x_i,y_i)) }{ \sum_{y \in \mathcal{Y}_i} \exp(\score_\theta(x_i,y) ) }
\end{align*}
where $\mathcal{Y}_i \subset \mathcal{Y}$ satisfying $y_i \in \mathcal{Y}_i$ is a set containing the gold and negative targets for the $i$-th labeled example.
We pre-encode every $y \in \mathcal{Y}$ to $v_y = \enc^\theta_Y(\mathrm{text}_Y(y))$ at test time and efficiently compute
the highest scoring target $\hat{y}(x) = \argmax_{y \in \mathcal{Y}}\; \inner{\enc^\theta_X(\mathrm{text}_X(x))}{v_y}$ for any $x \in \mathcal{X}$ by
maximum inner product search.

In multitask retrieval, there are $K$ retrieval tasks each with $N_k$ training examples
$(x^{(k)}_1,y^{(k)}_1) \ldots (x^{(k)}_{N_k},y^{(k)}_{N_k})\sim \pop_k$ drawn iid from the $k$-th population distribution $\pop_k$.
We use the KILT benchmark which includes $K=8$ tasks addressing QA, entity linking, fact checking, slot filling, and dialogue.\footnote{We write ``task'' and ``dataset'' synonymously
instead of distinguishing datasets from task types as done in some previous works. Thus KILT has 8 tasks and 5 task types.}
The per-task loss is
\begin{align*}
  J_k(\theta) =  -\frac{1}{N_k} \sum_{i=1}^{N_k} \log \frac{ \exp(\score_\theta(x^{(k)}_i,y^{(k)}_i)) }{ \sum_{y \in \mathcal{Y}^{(k)}_i} \exp(\score_\theta(x^{(k)}_i,y)) }
\end{align*}
defining the final loss
\begin{align*}
  J(\theta) =  \sum_{k=1}^K \frac{N_k}{N} \times J_k(\theta)
\end{align*}
Previous work by \citet{maillard2021multi} use the following setting.
The KB $\mathcal{Y}$ consists of 100-token disjoint Wikipedia passages.
The text mappings $\mathrm{text}_X,\mathrm{text}_Y$ apply the BERT tokenizer to unmodified queries and passages.
The encoders $\enc^\theta_X, \enc^\theta_Y$ are initialized with independent pretrained BERT-bases (uncased).
The task-specific training datasets are downsampled to be of similar sizes.
As in DPR, they train the model using hard negatives based on BM25, followed by one round of hard negative mining from the model
(only on Natural Questions and TriviaQA in which verifying if a candidate negative is indeed incorrect is expedient).

We now describe the main sources of improvement that we achieve over the baseline multitask retriever:
a better choice of the base model with appropriate prompting, and better optimization.

\subsection{Base Model}

We use a shared T5 to parameterize and initialize the query and passage encoder $\enc^\theta = \enc^\theta_X = \enc^\theta_Y$.
Specifically, we follow the ST5-EncDec architecture \citep{ni2021sentence} which encode any $z \in \mathcal{V}^+$ as
\begin{align*}
  \enc^\theta(z) = \mathrm{T5}.\mathrm{generate}(z,\; \mathrm{length}=1).\mathrm{state}
\end{align*}
(i.e., we run the T5-encoder on $z$, run the T5-decoder for 1 step from the special start symbol, and take the resulting hidden state prior to token prediction).
In addition, we define the text mapping for queries $x \in \mathcal{X}$ in task $k$ as
\begin{align*}
  \mathrm{text}_X(x) = \mathrm{T5Tokenizer}(\pi_k \oplus \text{[SEP]} \oplus x)
\end{align*}
where $\oplus$ is string concatenation, [SEP] is the special separation token, and $\pi_k$ is a text prefix that indicates which task $x$ is a query of.
We use dataset names as prefixes (e.g., $\pi_1 = $``NQ'').
The text mapping for passages $y \in \mathcal{Y}$ does not use prefixes, that is
\begin{align*}
  \mathrm{text}_Y(y) = \mathrm{T5Tokenizer}(y)
\end{align*}
This allows us to pre-encode passage embeddings at test time and retain the efficiency of the single-task dual encoder framework.

While simple, this choice is the most crucial component in our apporach to improving multitask retrieval.
We take a model pretrained for multitasking and adopt the same prefix concatenation scheme for task adaptation,
treating multitask retrieval as a continuation of the T5 training.

Interestingly, using task markers is reported to be not helpful in \citet{maillard2021multi}.
This is likely because their base model, BERT, is not pretrained for multitasking.
Another difference is that they use task markers to indicate the 5 task types (e.g., ``QA''),
whereas we use fine-grained markers to indicate the 8 tasks (e.g., ``NQ'').
While there are previous works that use T5 for dense retrieval \citep{ni2021sentence},
we are the first to exploit the multitasking component of T5 pretraining for multitask retrieval.

\subsection{Adaptive Learning}\label{sec:adaptive}

For the $k$-th task, the linear approximation of $J_k(\theta)$ around $a \in \R^d$ is
\begin{align*}
  J_k(\theta) \approx J_k(a) + \inner{\nabla J_k(a)}{\theta - a}
\end{align*}
Let $\theta^{(t)}$ denote the parameter value at the $t$-th update in gradient-based training.
For any $i = 1 \ldots d$, we define $\theta^{(t)}_{-i}$ to be equal to $\theta^{(t)}$ except that its $i$-th element is zero.
The approximation of $J_k(\theta)$ around $a = \theta^{(t)}_{-i}$ at $\theta = \theta^{(t)}$ is
\begin{align*}
  J_k(\theta^{(t)}) &\approx J_k(\theta^{(t)}_{-i}) + \inner{\nabla J_k(\theta^{(t)}_{-i})}{\theta^{(t)} - \theta^{(t)}_{-i}} \\
  &= J_k(\theta^{(t)}_{-i}) +  \frac{\partial J_k(\theta^{(t)})}{\partial \theta_i} \times \theta^{(t)}_i
\end{align*}
Rearranging and taking the absolute value, we have
\begin{align}
  \si_{i,k}^{(t)} = \abs{ \frac{\partial J_k(\theta^{(t)})}{\partial \theta_i} \times \theta^{(t)}_i } \approx   \abs{J_k(\theta^{(t)}) - J_k(\theta^{(t)}_{-i})} \label{eq:si}
\end{align}
which is easily computable and can be viewed as measuring how sensitive the $i$-th parameter is with respect to the $k$-th task in the $t$-th iteration of training.
We propose to use this quantity, previously used in the model pruning literature \citep{molchanov2016pruning},
to encourage task specialization during training.
We define a conditional distribution over $K$ tasks by
\begin{align}
  q(k|\theta^{(t)}, t, i) = \frac{\exp(\bar{\si}_{i,k}^{(t)} / \tau_t)}{\sum_{k=1}^K \exp(\bar{\si}_{i,k}^{(t)} / \tau_t)} \label{eq:dist}
\end{align}
where $\tau_t > 0$ is a temperature and $\bar{\si}_{i,k}^{(t)}$ is an appropriately normalized and amortized estimation of $\si_{i,k}^{(t)}$ in Eq.~\eqref{eq:si} (see Section~\ref{sec:sensitivity}).
Assuming training examples are sampled to roughly balance the size across tasks (i.e., $N_k \approx N_{k'}$),
we take the following gradient step for the $i$-th parameter in the $t$-th iteration:
\begin{align*}
  \theta^{(t+1)}_i = \theta^{(t)}_i - \eta \sum_{k=1}^K q(k|\theta^{(t)}, t, i) \times \frac{\partial J_k(\theta^{(t)})}{\partial \theta_i}
\end{align*}
Note that this is a per-parameter adaptive learning.
Each parameter $\theta_i \in \R$ maintains a distribution over $K$ tasks and is updated more aggresively for tasks that $\theta_i$ is sensitive to.

\subsubsection{Sensitivity normalization}
\label{sec:sensitivity}
The $d$ parameters $\theta^{(t)}$ can be of very different magnitudes.
To reduce the parameter-wise variance in the sensitivity scores, for task $k$ we divide the scores by the median of across all parameters with respect to task $k$:
\begin{align*}
  \tilde{\si}_{i,k}^{(t)} = \frac{\si_{i,k}^{(t)}}{\mathrm{median}_{j=1 \ldots d}(\si_{j,k}^{(t)})}
\end{align*}
We use the median instead of the mean to account for the long tail distribution of task-specific sensitivity scores.
We also use momentum to amortize the scores: assuming some $\be > 0$
\begin{align*}
  \bar{\si}_{i,k}^{(t)} = (1-\be) \bar{\si}_{i,k}^{(t-1)} + \be \tilde{\si}_{i,k}^{(t)}
\end{align*}
where $\bar{\si}_{i,k}^{(0)} = 0$.
This is the final version of sensitivity that we use in Eq.~\eqref{eq:dist}.
The algorithm in matrix form is given in Algorithm~\ref{alg:taco} (Appendix~\ref{app:alg}).

\section{Experiments}

\subsection{Setup}

\paragraph{Datasets.}
We follow \cite{maillard2021multi} and use eight tasks from KILT \citep{petroni2021kilt} for training and evaluation. We randomly downsample the training data of the two largest datasets (T-REx and zsRE) to the same order of magnitude as the rest. All the datasets share the same knowledge base of 36 million disjoint 100-token Wikipedia passages preprocessed by \citet{maillard2021multi}. The data statistics and other data-related details can be found in
Appendix~\ref{app:data_details}.

\paragraph{Evaluation.}
We use the page-level $R$-precision (the suggested main metric in KILT) to measure the retrieval performance. Page-level $R$-precision is the fraction of the $R$ gold pages captured by the retriever in the top-$R$ candidates. We map the retrieved passages to the their corresponding pages and use official KILT evaluation scripts to evaluate the page-level $R$-precision. We also report passage-level $R$-precision proposed by \citet{maillard2021multi} on dev sets in Appendix~\ref{app:passage_level}. We use TREC Eval\footnote{\url{https://trec.nist.gov/trec_eval/}} to evaluate the passage-level $R$-precision.

\paragraph{Model details.}
We initialize our dual encoder with the official T5-base \citep{raffel2019t5} checkpoint. The query encoder and passage encoder share weights. Following the ANCE \citep{xiong2020approximate} training paradigm, we first warmup our model for 20 epochs with BM25 hard negatives by naive multitask learning with task prefix. Then we train the model for 8 ANCE episodes with the model-mined hard negatives refreshed at the begining of each ANCE episode. We adopt naive multitask learning with task prefix for the first 7 ANCE episodes and apply the adaptive learning introduced in Section~\ref{sec:adaptive} for the last ANCE episode to improve the performance further. We use Adam \citep{adam14} with  a linear learning rate decay schedule with warmup proportion 0.1 over 3 epochs for each ANCE iteration.
We provide more details and hyperparameters in Appendix~\ref{app:other_details}.

\begin{table*}[t]
  \setlength{\tabcolsep}{3.4pt}
  \renewcommand{\arraystretch}{1.2}
  \small
  {\centering
    \begin{tabular}{lcccccccc|c}
      \Xhline{2\arrayrulewidth}
      \rule{0pt}{1.1em}& Fact Check. & Ent. L. & \multicolumn{2}{c}{Slot Filling}  & \multicolumn{3}{c}{Open Domain QA} & Dial. & \\
       \rule{0pt}{1.1em}{\bf Model} & {\bf FEV} & {\bf AY2} & {\bf T-REx} & {\bf zsRE} & {\bf NQ} & {\bf HoPo} & {\bf TQA} & {\bf WoW} & {\bf Avg} \\
           \hline\\[-1em]
         \multicolumn{10}{c}{\textbf{Baselines.
         }} \\
         BM25$^*$ & 50.13  & 3.47 & 58.60  & 66.43  & 25.83 & 43.95  & 29.44  & 27.50  & 38.17  \\
               BART$_{mt}^\dagger$ & 81.92  &  89.17 &  75.18  & 91.08  & 58.62  & 48.69  & 67.64  & 50.98  & 70.41 \\
          CorpusBrain$_{mt}^\dagger$ & \underline{82.06}   &  {\bf 90.84} & \underline{77.62}  & 98.26  & 59.10  & 50.07  & 68.78  & \underline{53.75}  & \underline{72.56}  \\
       MT-DPR$^*$  &  74.72  & 83.78 & 69.18  & 77.23  & 61.51 & 44.21  & 61.95  & 39.70  & 64.04  \\
       Task-specific DPR$^*$ &  73.60  & 81.77 & 69.08  & 97.74  & \underline{63.24}  & 46.63  & 65.12  & 40.32  & 67.19 \\
       Task-specific BART$^\dagger$ & 80.03 & 87.98 & 74.46 & 93.91 & 50.96 & 39.21 & 66.13 & 50.75 & 67.93 \\
       Task-specific CorpusBrain$^\dagger$ & 81.77 & \underline{90.36} & 76.90 & \underline{98.49} & 57.67 & {\bf 50.62} & \underline{69.25} & 53.60 & 72.33 \\
       Task-specific (ours)  &  74.28  & 85.28 & 77.18  & {\bf 99.38} & {\bf 65.39}  & 46.79  & 69.08  & 53.63  & 71.38  \\
         \hline
       \multicolumn{10}{c}{\textbf{Non-Comparable Models (For Reference).}} \\
       CorpusBrain$_{mt+BLINK}^\dagger$ & 85.03 & 92.86 & 80.22 & 98.49 & 64.61 & 52.23 & 71.71 & 59.72 & 75.61 \\
       GENRE$^\dagger$  & 84.68 & 92.75 & 79.68 & 94.84 & 64.26 & 51.82 & 71.11 & 56.32 & 74.43 \\
       TABi \citep{leszczynski2022tabi} & 85.8 & - & 82.0 & 95.2 & 62.4 & 52.7 & 71.5 & 51.8 & - \\
   \hline
        TACO  & {\bf 86.17}  & 84.64  & {\bf 78.12}  & 97.91  &  61.86  & \underline{50.61}  & {\bf 69.62}  & {\bf 60.97}  & {\bf 73.74}  \\
      \Xhline{2\arrayrulewidth}
    \end{tabular}
    \caption{Page-level R-precision on KILT validation data.
    {\bf Bold} indicates the best  model and \underline{underline} indicates the second. $\dagger$ and $*$ mark results from \citet{chen2022corpusbrain} and \citet{maillard2021multi} respectively. The non-comparable models are trained on additional data or use extra information. We list them only for reference not for comparison. Taks-specific models use a separate retriever for each task while all the other models use a single retriever across all the tasks.}
  \label{tab:main_dev}
  }
\end{table*}

\begin{table*}[hbt!]
  \setlength{\tabcolsep}{4.2pt}
  \renewcommand{\arraystretch}{1.2}
  \small
  {\centering
    \begin{tabular}{lcccccccc|c}
      \Xhline{2\arrayrulewidth}
      \rule{0pt}{1.1em}& Fact Check. & Ent. L. & \multicolumn{2}{c}{Slot Filling}  & \multicolumn{3}{c}{Open Domain QA} & Dial. & \\
       \rule{0pt}{1.1em}{\bf Model} & {\bf FEV} & {\bf AY2} & {\bf T-REx} & {\bf zsRE} & {\bf NQ} & {\bf HoPo} & {\bf TQA} & {\bf WoW} & {\bf Avg} \\
           \hline\\[-1em]
         \multicolumn{10}{c}{\textbf{Baselines.
         }} \\
         TF-IDF$^\dagger$ & 50.9  & 3.7 & 44.7  & 60.8  & 28.1 & 34.1  & 46.4  & 49.0  & 39.7  \\
               SEAL$^\ddagger$ & \underline{81.4}  &  - &  62.1  & 91.6  & {\bf 63.2}  & {\bf 58.8}  & 68.4  & \underline{57.5}  & - \\
       MT-DPR$^*$  &  74.5  & 26.5 & 69.5  & 80.9  & 59.4 & 42.9  & 61.5  & 41.1  & 57.0  \\
       MT-DPR$_{WEB}^\ddagger$ &  74.8  & - & 75.6  & 89.7  & 59.8  & 45.4  & 58.9  & 41.5  & - \\
       Task-specific (ours)  &  73.22  & \underline{79.52} & \underline{77.00}  & {\bf 99.15} & \underline{60.87}  & 46.50  & {\bf 69.12}  & 55.03  & 70.05  \\
         \hline
       \multicolumn{10}{c}{\textbf{Non-Comparable Models (For Reference).}} \\
       CorpusBrain$_{mt+BLINK}^\dagger$ & 84.07 & 89.98 & 79.98 & 98.27 & 60.32 & 51.80 & 70.19 & 64.79 & 74.93 \\
       GENRE$^\dagger$  & 83.64 & 89.85 & 79.42 & 95.81 & 60.25 & 51.27 & 69.16 & 62.88 & 74.04 \\
       TABi \cite{leszczynski2022tabi} & 84.4 & - & 81.9 & 96.2 & 62.6 & 53.1 & 70.4 & 59.1 & - \\
   \hline
        TACO  & {\bf 84.07}  & {\bf 80.64}  & {\bf 77.22}  & \underline{98.21}  &  60.80  & \underline{50.70}  & \underline{68.45}  & {\bf 62.64}  & {\bf 72.84}  \\
      \Xhline{2\arrayrulewidth}
    \end{tabular}
    \caption{Page-level R-precision on KILT test data.
    {\bf Bold} indicates the best  model and \underline{underline} indicates the second. $\dagger$, $*$ and $\ddagger$ mark results from \citet{chen2022corpusbrain}, \citet{maillard2021multi} and \citet{bevilacqua2022autoregressive} respectively. The non-comparable models are trained on additional data or use extra information. We list them only for reference not for comparison.}
  \label{tab:main_test}
  }
\end{table*}

\subsection{Main Results}
We refer to our model as \textbf{TACO}, which stands for \textbf{TA}sk spe\textbf{C}ialty \textbf{O}ptimization.
Table~\ref{tab:main_dev} and Table~\ref{tab:main_test} show our main results on the KILT validation data and test data respectively. Fewer comparable baselines are available for KILT test data than for KILT validation data.

\begin{table*}[hbt!]
  \setlength{\tabcolsep}{4.2pt}
  \renewcommand{\arraystretch}{1.2}
  \label{tab:main_ab}
  \small
  \begin{center}
    \begin{tabular}{lcccccccc|c}
      \Xhline{2\arrayrulewidth}
      \rule{0pt}{1.1em}& Fact Check. & Ent. L. & \multicolumn{2}{c}{Slot Filling}  & \multicolumn{3}{c}{Open Domain QA} & Dial. & \\
       \rule{0pt}{1.1em}{\bf Variants} & {\bf FEV} & {\bf AY2} & {\bf T-REx} & {\bf zsRE} & {\bf NQ} & {\bf HoPo} & {\bf TQA} & {\bf WoW} & {\bf Avg} \\
           \hline\\[-1em]
    TACO & {\bf 86.17}  & 84.64  & {\bf 78.12}  & {\bf 97.91}  &  61.86  & 50.61  & {\bf 69.62}  & {\bf 60.97}  & {\bf 73.74}  \\
     w/o task prefix &  85.71 & 84.68  & 74.82 & 94.68 & 61.05 & 49.38 & 67.79 & 58.81 & 72.12 \\
     w/o adaptive & 84.81 & 85.49 & 75.00 & 92.24 & 62.81 & 51.47 & 68.95 & 60.54 & 72.66\\
     w/o task prefix w/o adaptive  &  84.03  & 85.62 & 70.96 & 86.04  & 62.46  & 49.78  & 66.04  & 59.95  & 70.61  \\
      task query encoder & 82.71   & {\bf 87.56} & 72.72 & 85.15 & {\bf 64.01} & 49.74 & 69.12 & 55.93 & 70.87 \\
      task type marker &  84.49 & 85.51  & 73.88 & 89.37 & 62.85 & 50.97 & 67.70 & 60.02 & 71.85 \\
      PCG \cite{yu2020gradient} &  84.97  & 85.26 & 74.90 & 91.43  & 62.67  & 51.47  & 68.54  & 60.48  & 72.47  \\
      CGD \cite{piratla2021focus} &  82.25  & 80.39 & 71.62 & 83.40  & 62.67  & 49.66  & 66.73  & 59.33  & 69.51  \\
      GradNorm \cite{chen2018gradnorm} &  84.70  & 85.28 & 75.32 & 91.73  & 63.80  & {\bf 51.97}  & 69.30  & 60.31  & 72.80  \\
      \Xhline{2\arrayrulewidth}
    \end{tabular}\caption{Ablation study results on KILT validation data. We report page-level R-precision. {\bf Bold} indicates the best variant. Each line makes a single or multiple changes from the TACO model. The performance of the recent general multitask algorithms, PCG \cite{yu2020gradient}, CGD \cite{piratla2021focus} and GradNorm \cite{chen2018gradnorm}, are obtained from our own implementation.
  }\label{tab:ablation}
    \end{center}
\end{table*}

Let avg val denote average validation page-level R-Precision. TACO achieves the best performance on 4 out of 8 tasks for both validation and test data. The performance is either the second best or close to  the second best except AIDA, an entity linking dataset favoring autoregressive retrieval models over dense retrieval models \citep{cao2021autoregressive}. TACO outperforms the previous multitask dense retrieval model MT-DPR \citep{maillard2021multi} significantly (+7.34\% avg val). TACO also achieves better performance compared with current top performing multitask autoregressive retrieval models in comparable setting (finetuned purely on KILT). TACO outperforms BART$_{mt}$ (+3.33\% avg val) with smaller model size (T5-base vs Bart-large). Compared with BART$_{mt}$, CorpusBrain$_{mt}$ employs additional pretraining and yields significant improvement over BART$_{mt}$ (+2.15\% avg val).
TACO still outperforms CorpusBrain$_{mt}$ (+1.18\% avg val) with smaller model size and no additional pretraining. We also list various top performing multitask retrieval models for reference but not for comparison because they are not in comparable setting. Both GENRE and CorpusBrain$_{mt+BLINK}$ are finetuned on a large amount of additional training data besides KILT training data. Specifically, they also use BLINK training data \citep{wu2019zero} for finetuning, which contains 8.9M annotated wikipedia sentences. TABi \citep{leszczynski2022tabi} uses extra type labels information and leverages knowledge graph that is very effective for retrieval. TACO even rivals these non-comparable models on all the tasks except AIDA.

TACO is the only model that outperforms strong task-specific models noticeably. Our task-specific baseline is significantly stronger than the task-specific DPR, likely due to  better training paradigm (ANCE) and  better model (T5 vs BERT). Task-specific CorpusBrain is even stronger, especially for FEVER and AIDA. Only TACO and CorpusBrain$_{mt}$ outperform the strong task-specific models. TACO achieves a 2.36\% improvement over its task-specific counterpart and a 1.41\% improvement over the task-specific CorpusBrain, but CorpusBrain$_{mt}$ is only slightly better than its task-specific counterpart (+0.23\% avg val).

\subsection{Analysis}

\subsubsection{Ablation Study}
Table~\ref{tab:ablation} shows the results of ablation studies on KILT validation data.
\paragraph{Model components.}
We first conduct experiments to understand the impact of individual components of our model.  Removing task prefix results in 1.62\% R-precision decrease and disabling adaptive learning yields 1.08\% R-precision decrease. Removing both task prefix and adaptive learning significantly degrades the performance (-3.13\%). This demonstrates that both task prefix and adaptive learning contribute to the effectiveness of TACO.
\paragraph{Query variants.} We conduct experiments to investigate other query side variants besides task prefix. These variants are not trained with adaptive learning and only change the query input format or model. Leveraging task-specific query encoder yields slightly better performance (70.87\% vs 70.61\%), but is outperformed by task prefix significantly (70.87\% vs 72.66\%). The task type marker introduced in \citet{maillard2021multi} is not helpful for BERT-based MT-DPR, but we find them effective for our T5-based model. This is likely because T5 is pretrained for multitasking. We conduct experiments to leverage their task type markers for our model. Using task type markers (i.e., 5 symbols indicating the 5 classes of task in KILT) leads to 1.24\% R-precision improvement (71.85\% vs 70.61\%), but is less effective than our fine-grained dataset-level task prefix (71.85\% vs 72.66\%).

\paragraph{Mutltitask learning variants.} We compare our adaptive learning method with recent general multitask learning algorithms with our own implementation. PCG \citep{yu2020gradient} focuses on mitigating the conflict of gradients from different tasks. It performs on par with the ``w/o adaptive'' variant (72.47\% vs 72.66\%), but underperforms TACO which leverages our adaptive learning (72.47\% vs 73.74\%). This shows that the gradient conflict is not the main bottleneck in our multitask retrieval setting. CGD \citep{piratla2021focus} aims to improve multitask learning by encouraging update towards common directions of different tasks, which is opposite to our method that encourages task specialties. It performs much worse than TACO (69.51\% vs 73.74\% and lags behind the ``w/o adaptive'' variant significantly (69.51\% vs 72.66\%). This shows that we should encourage task specialty rather than emphasizing tasks shared part for multitask retrieval.
GradNorm \cite{chen2018gradnorm} tries to weight different tasks losses by using the average gradient norm. It performs slightly better than the naive ``w/o adaptive'' variant (72.47\% vs 72.66\%).  Our adaptive learning method achieves descent improvement over GradNorm (73.74\% vs 72.80\%). Note that our adaptive update is more fine-grained and critically different because we adjust learning rates along both task dimension and parameter dimension compared with GradNorm that only do loss re-weighting.

\begin{figure}[t!]
  \centering

  \begin{subfigure}{0.45\columnwidth}
    \centering
    \includegraphics[width=\linewidth]{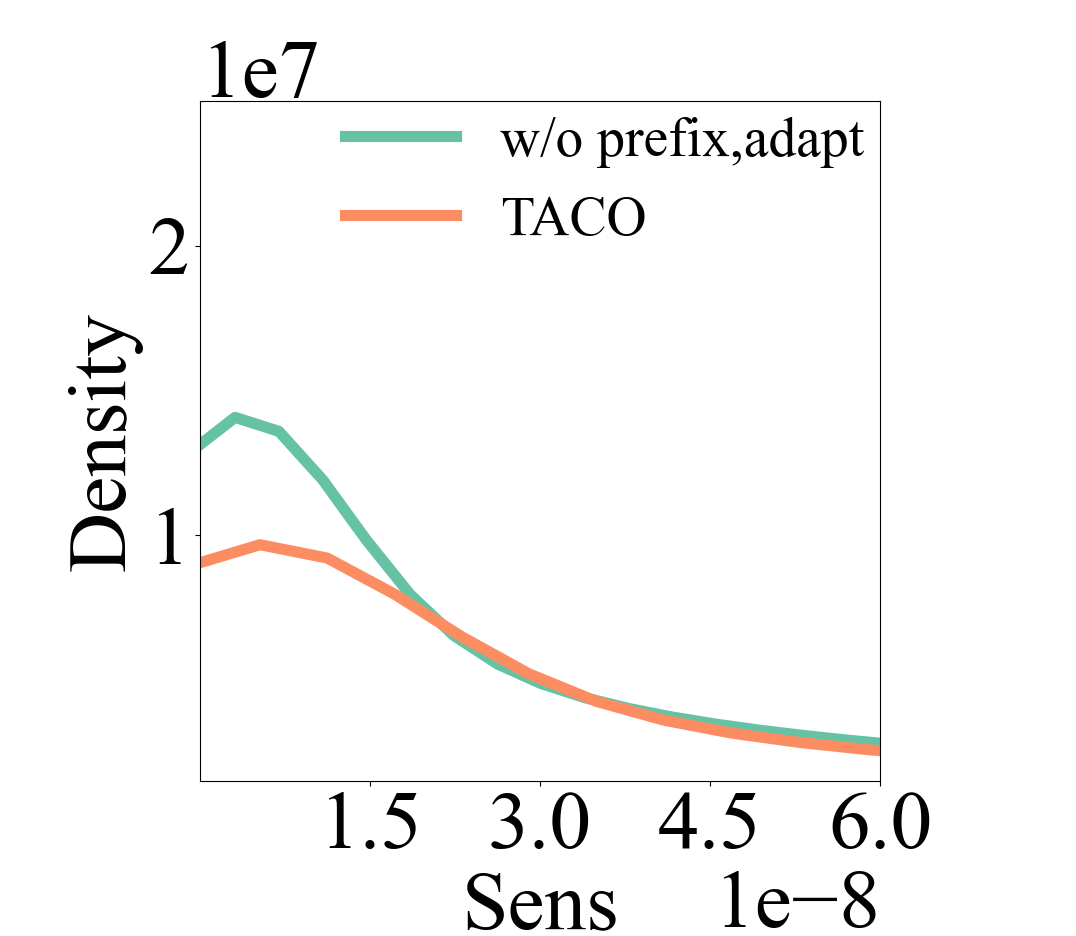}
    \caption{NQ}
    \label{fig:nq_dist}
  \end{subfigure}
  \hfill
  \begin{subfigure}{0.45\columnwidth}
    \centering
    \includegraphics[width=\linewidth]{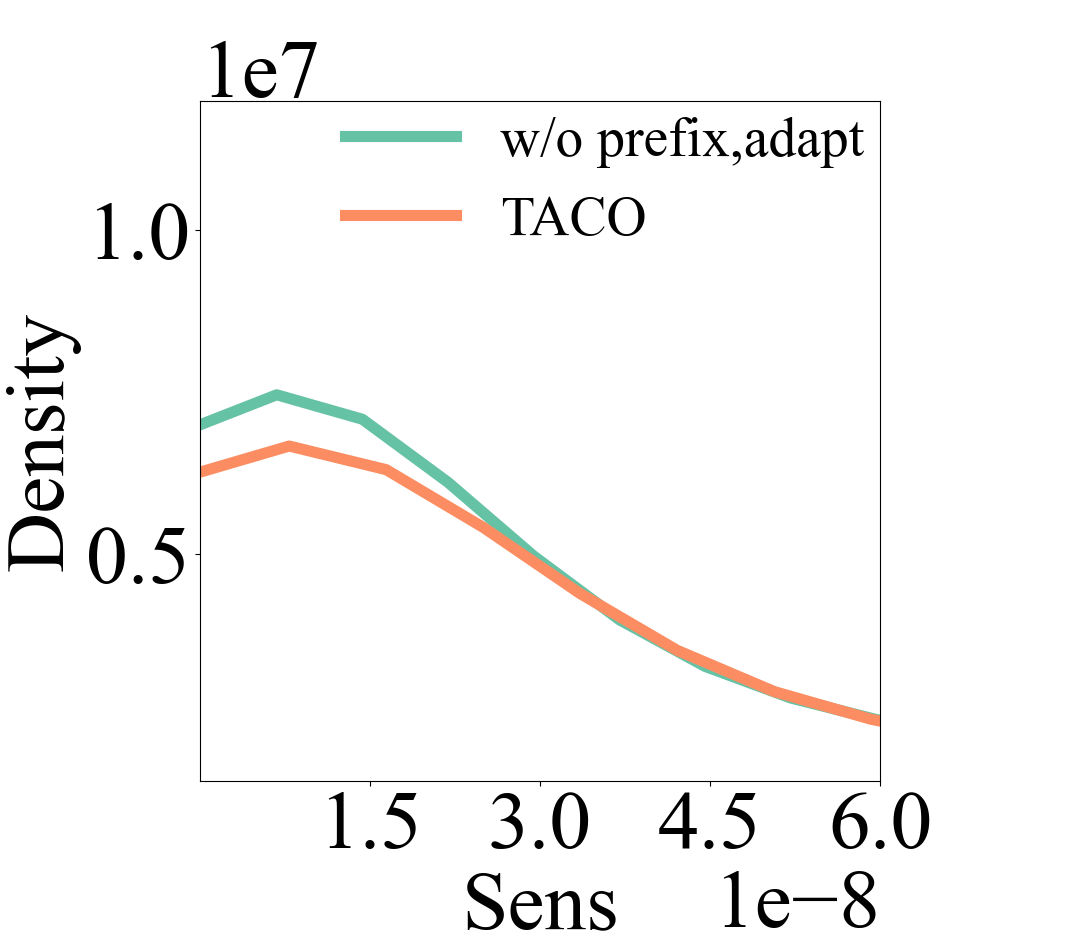}
    \caption{WoW}
    \label{fig:wow_dist}
  \end{subfigure}

  \vspace{0.5cm}

  \begin{subfigure}{0.45\columnwidth}
    \centering
    \includegraphics[width=\linewidth]{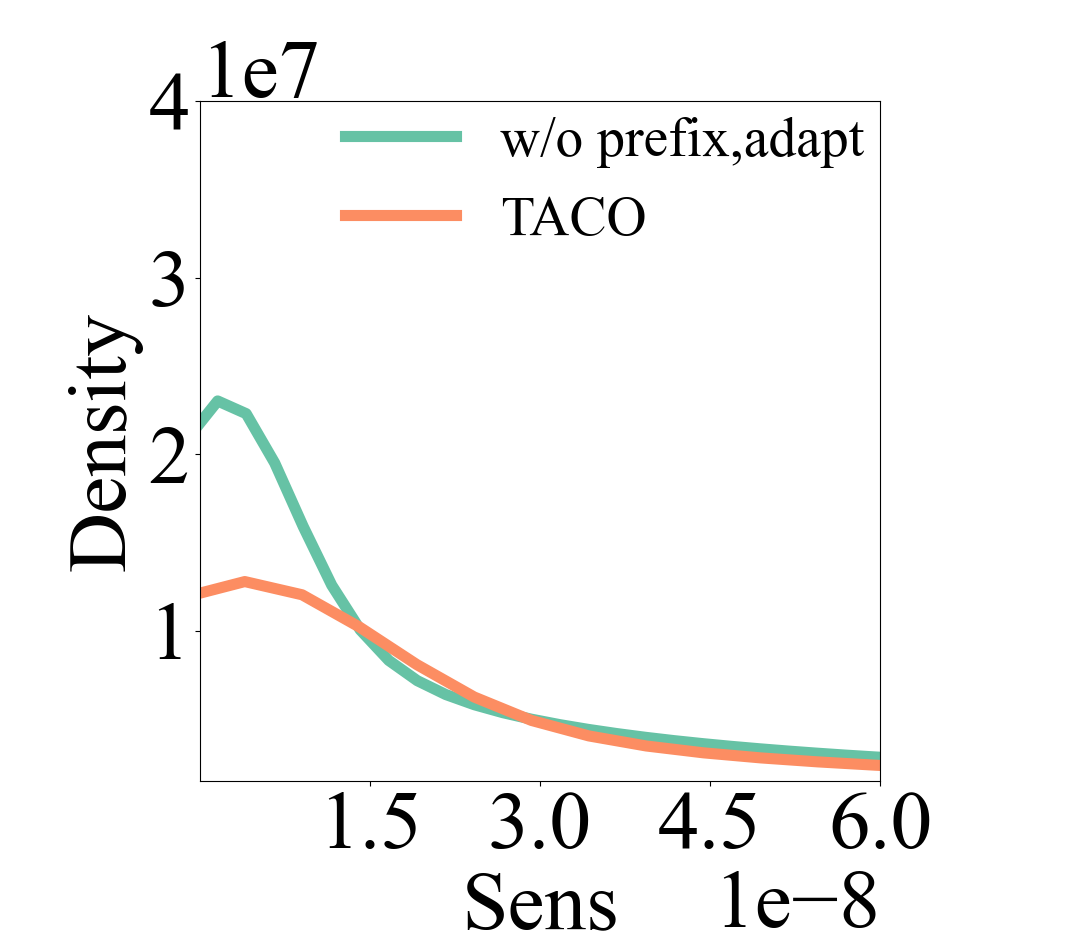}
    \caption{T-REx}
    \label{fig:trex_dist}
  \end{subfigure}
  \hfill
  \begin{subfigure}{0.45\columnwidth}
    \centering
    \includegraphics[width=\linewidth]{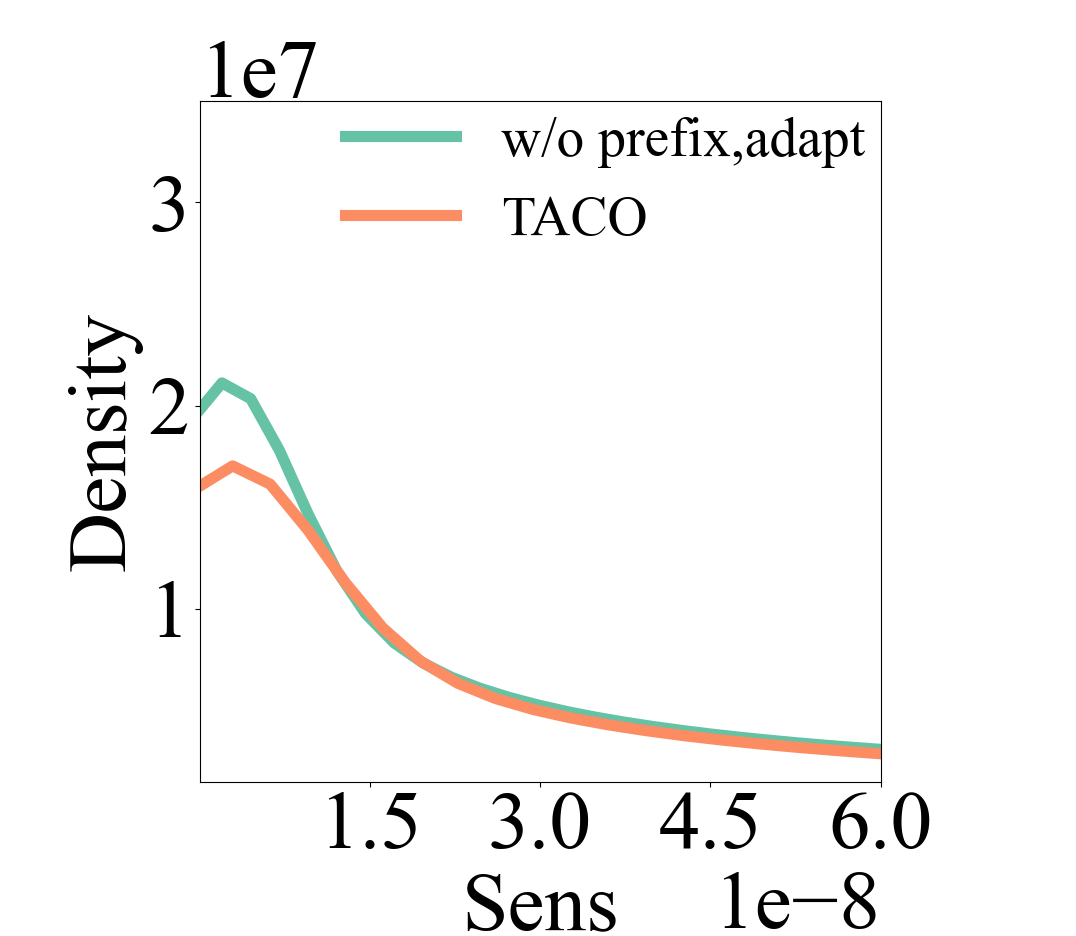}
    \caption{FEV}
    \label{fig:fev_dist}
  \end{subfigure}

  \caption{Task-specific sensitivity density distribution on the training data of four KILT tasks. The final models are used.  The x-axis is sensitivity, and we drop outliers that are far from the median to ease visualization.}
  \label{fig:sens}
\end{figure}

\paragraph{Adaptive learning.} We consider variations of the main version of adaptive learning which is applied only in the last ANCE episode. Specifically, we investigate the impact of applying adaptive learning to the last four ANCE episodes using an exponential softmax temperature decay scheduler. This approach yields an average page-level R-precision of 73.47\%. In comparison, when adaptive learning is applied only to the last ANCE episode, we achieve an average page-level R-precision of 73.74\%. These results suggest that extending adaptive learning to more ANCE episodes does not yield improvement.
Additionally, we examine the effectiveness of encouraging task specialization within adaptive learning. For this purpose, we focus on the second ANCE episode and experiment with positive softmax temperature (encouraging task specialty) and negative softmax temperature (discouraging task specialty). Encouraging task specialization results in an average page-level R-precision of 70.53\%, while discouraging task specialization leads to an average page-level R-precision of 68.39\%. In comparison, the performance of the standard multitask baseline at the second ANCE episode is 69.28\%. These results highlight the benefits of encouraging task specialization and the detrimental effect of discouraging task specialization within adaptive learning.
Normalizing task sensitivity using the median is preferred over using the mean or not applying any normalization, as different tasks exhibit variations in magnitude while sharing similar distribution shapes (see Figure~\ref{fig:sens}).

\begin{figure}[t!]
\begin{center}
    \includegraphics[width=\columnwidth]{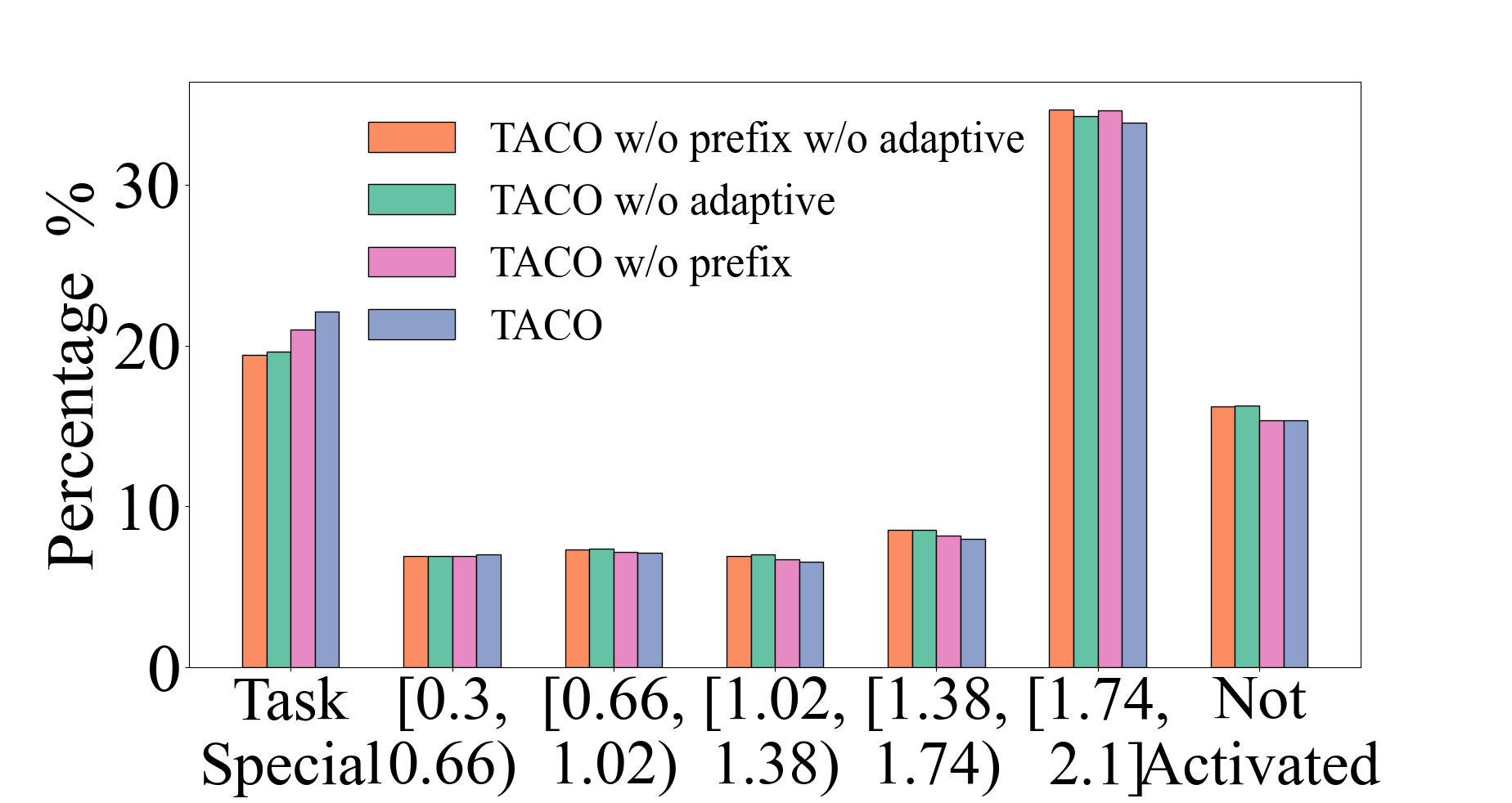}
    \caption{Task entropy histograms for model variants}\label{fig:entropy}
\end{center}
\end{figure}

\subsubsection{Task Specialization}
Figure~\ref{fig:entropy} plots the histograms of task entropy for the learned parameters. The task entropy for each parameter is calculated with the distribution defined in equation~\ref{eq:dist}. We first group parameters into two special bins. The first is  a ``Task Specific'' bin that includes parameters whose entropy is smaller than 0.3, which is the entropy of 95\% probability on one task and the 5\% uniformly on the rest seven.
The ``Not Activated'' bin includes parameters whose sensitivity w.r.t. all tasks is near zero ($<1e-8$).
TACO significantly improves the fraction of task specific parameters to 22\%, in comparison with 19\% in naive multitask model (w/o prefix w/o adaptive). It also reduces the fraction of not activated parameters, showing optimizing task specialty also better utilizes the model capacity.

Figure~\ref{fig:sens} plots the kernel density estimated distribution of task-specific sensitivity in TACO and the standard multitask model for four KILT tasks. We drop outliers that deviates significantly from the median to ease visualization.  Notably, TACO exhibits a noticeable reduction in the peak on the low sensitivity side for each task compared to the standard multitasking model. This observation suggests that TACO activates a larger number of parameters and enhances their sensitivity towards individual tasks.

\subsubsection{Additional Benchmark}

To test the performance of TACO in a different setup other than KILT, we constructed an additional benchmark containing MS-MARCO \citep{nguyen2016ms}, ZESHEL \citep{logeswaran2019zero}, a document-level version of FEVER from BEIR \citep{thakur2021beir}, and Natural Questions from KILT.
We chose this combination for a few reasons.
First, we found that few public datasets outside KILT provide sufficiently large and high-quality training data other than MS-MARCO and ZESHEL.
Second, each task now has its own KB to retrieve from,
making this a rather different setup from KILT in which all tasks share one KB.
We compare task-specific retrievers and multitask retrievers trained by TACO and
other methods.
Table~\ref{tab:benchmark2} shows their recall at 100 on the validation split.
We see that multitasking is clearly beneficial for this benchmark.
The best performance is obtained by CGD and it is the only multitask optimization method that yields noticeable improvements over the standard multitask model.  Given that CGD aims to improve
multitask learning by encouraging update towards
common directions of different tasks,
we hypothesize that the need for task specialization is diminished here because the tasks are more similar in difficulty (e.g., in KILT, T-REx and zsRE are much easier than HotpotQA).
This experiment sheds light on what multitask settings most benefit from task specialization.

\begin{table}[t]
  \setlength{\tabcolsep}{4.2pt}
  \renewcommand{\arraystretch}{1.2}
  \label{tab:main_ab}
  \small
  \begin{center}
    \begin{tabular}{lcccc|c}
      \Xhline{2\arrayrulewidth}
        & {\bf MS} & {\bf ZES} & {\bf FEV} & {\bf NQ} & {\bf Avg} \\
           \hline\\[-1em]
    Task-specific  & 73.3 & 67.3 & 90.0 & 71.8 & 75.6 \\
    TACO & 85.8 & 67.6 & 91.2 & 76.8 & 80.4 \\
    w/o adapt  & 85.9 & 67.5 & 91.3 &76.6 & 80.3 \\
    w/o prefix, adapt  & 86.2 & 68.1 & 92.2 & 76.4 & 80.7 \\
    PCG & 86.1 & 67.9 & 91.7 & 76.9 & 80.7 \\
    CGD & 86.8 & 69.1 & 94.4 & 76.2 & 81.6 \\
    GradNorm & 86.0 & 67.3 & 91.7 & 76.9 & 80.5 \\
      \Xhline{2\arrayrulewidth}
    \end{tabular}\caption{Recall@100 on an additional benchmark
    containing MS-MARCO (MS), ZESHEL (ZES), FEVER (FEV), and Natural Questions (NQ).
  }\label{tab:benchmark2}
    \end{center}
\end{table}

\section{Conclusions}

Multitask retrieval has compelling practical advantages such as model simplicity and memory efficiency, but it lags behind task-specific retrieval in the existing literature. We have shown that it is possible to significantly improve the performance of multitask retrieval by promoting task specialization. The key steps are the use of a base model optimized for multitasking with appropriate prompting and a per-parameter adaptive learning technique that upweights the task gradients by the parameters' sensitivity to the task losses. We have achieved strong results on the KILT retrieval benchmark.

\bibliography{tacl2021}
\bibliographystyle{acl_natbib}

\appendix

\section{Algorithm in Matrix Form}\label{app:alg}
Alogrithm~\ref{alg:taco} is the matrix form of our adaptive learning algorithm.

\begin{algorithm*}[t!]
    \caption{Task sensitivity-guided adaptive learning} \label{alg:taco}
    \begin{algorithmic}[1]
        \Require Model parameter $\theta \in \R^d$; minibatches $\mathcal{B}$ where each batch $B \in \mathcal{B}$ is further divided by tasks $B = \myset{B_k}_{k = 1\ldots K}$; moving average rate $\beta \in [0, 1]$; temperature $\tau > 0$; learning rate $\eta > 0$

        \Ensure  $\mathrm{median}: \R^{d \by K } \rightarrow \R^{K}$ is the column-wise median; $\mathrm{softmax}: \R^{d \by K} \rightarrow \R^{d \by K}$ is the row-wise softmax; $\textbf{1}_{K}$ is a vector of $K$ ones; $\odot$ is the Hadamard product.
        \State  Initialize $I \gets \mathbf{0} \in \R^{d \times K}$.
        \For{each batch $B = \myset{B_k}_{k = 1\ldots K}$ in $\mathcal{B}$}
              \State Compute the task-specific loss $J_k(\theta)$ on $B_k$ for each $k = 1 \ldots K$.
              \State Compute the gradient matrix $G \in \R^{d \times K}$ with each column $G_k \gets \nabla J_k(\theta)$.
              \State Compute the sensitivity matrix  $I' \in \R^{d \times K}$ with each column $I'_k \gets G_k \odot \theta$.
              \State Normalize the sensitivity scales across tasks  $I' \gets I' \diag{\mathrm{median}(I')}^{-1}$.
              \State Update the moving average $I \gets \beta I + (1-\beta) I'$.
              \State Update the parameter $\theta \gets \theta - \eta (G \odot U) \textbf{1}_{K}$ where $U \gets \mathrm{softmax}(I/\tau) \in \R^{d \by K}$.
        \EndFor

    \end{algorithmic}
\end{algorithm*}

\section{Data Details}\label{app:data_details}
See Table~\ref{tab:data_stats} for data statistics and some data-related hyperparameters. We randomly downsample T-REx and zsRE to bring them to the same order of magnitude as the others. We follow \cite{raffel2019t5} and use temperature-scaled mixing sampling strategy to compute  batch size for each task $k$: $B_k\propto  (N_k/\sum_{k'=1}^K N_{k'})^{1/c}$ for some temperature $c$ (we set it to 4 in our experiments). Here $N_k$ is the dataset size of task $k$. Note that we compute task loss of each task batch independently instead of mixing all task batches for every optimization step. Each dataset needs to sample different number of batches to cover every training sample in that dataset once. We set the maximum of them as the number of batches that every dataset needs to sample. We shuffle and cycle batch sampling iterators of datasets that finish iterating early. Batch size of each dataset computed by setting mixing temperature $c=4$ and $\sum_{k'=1}^K N_{k'}=120$ is in Table ~\ref{tab:data_stats}.
\begin{table}[t]
    \centering
    \begin{tabular}{l|ccc}
    \hline
    Dataset & \#Train & B & L\\
    \hline\\[-1em]
    Natural Questions    &  76k & 16 & 32\\
    TriviaQA     &  53k & 14 & 32 \\
    HotpotQA &  69k & 15 & 32\\
    Wizard of Wikipedia  &  80k & 16 & 256\\
    T-REx  & 95k & 16 & 32 \\
    FEVER  & 71k & 15 & 64\\
    Zero Shot RE  & 100k & 17 & 32\\
    AIDA-YAGO 2  & 18k & 11 & 128\\
    \hline
    \end{tabular}
    \caption{Data statistics and some data-related hyperparameters for our experiments. B denotes batch size. L denotes query maximum input length excluding the task prefix.}
    \label{tab:data_stats}
\end{table}

\section{Other Training Details}\label{app:other_details}
\begin{table}[h]
    \setlength{\tabcolsep}{2pt}
    \renewcommand{\arraystretch}{1.2}
    \begin{center}
    \begin{tabular}{ccccccc}
    \hline
      lr  & warmup  & \#negs & epochs & $\tau$ & $\beta$ & $B_{total}$ \\
      \hline\\[-1em]
      5e-6   &  0.1 & 2 & 3 & 2 & 0.999 & 120 \\
      \hline
    \end{tabular}
    \caption{Training hyperparameters for training our TACO-DR model. We use Adam\cite{adam14} with learning rate $5e-6$. We use linear learning rate schedule with warmup raio 0.1. Each query uses 2 hard negatives for training. Each ANCE episode trains for 3 epochs. Total batch size of all task batches are 120.}
    \label{tab:train_hyps}
    \end{center}
\end{table}

The data-related hyperparameters, such as maximum input query length and batch size, are listed in Table ~\ref{tab:data_stats}.
The training hyperparameters are listed in Table  ~\ref{tab:train_hyps}. We use NCE loss with cross device in-batch negative mixed with hard negatives to compute each task loss.  We sample two hard negatives for each query. We employ a ``burn in'' period for the first 10\% training steps with uniform learning rates for parameters to declare their tendency during adaptive learning.
All of our experiments are run on a machine with 8 A100-80GB GPUS. Our implementations are built upon OpenMatch~\citep{liu2021openmatch}.

\section{Softmax Temperature and Momentum Ratio}
\begin{table}[htb!]
\setlength{\tabcolsep}{2.2pt}
  \renewcommand{\arraystretch}{1.2}
  {\small
  \centering
\begin{tabular}{lcccccccc}
\hline
\textbf{$\tau$} & 0.1 & 1 & 2 & 5 & 10 & 100 \\
\hline\\[-1em]
\textbf{Avg R-prec}   & 72.45 & 73.23 & 73.74 & 73.72 & 73.48 & 72.85  \\
\hline
\end{tabular}
\caption{Average page-level R-precision w.r.t softmax temperature for our adaptive learning
}
\label{tab:temp}
}
\end{table}

\begin{table}[htb!]
\setlength{\tabcolsep}{2.2pt}
  \renewcommand{\arraystretch}{1.2}
  {\small
  \centering
\begin{tabular}{lcccccc}
\hline
\textbf{$\beta$} & 0 & 0.6 & 0.7 & 0.8 & 0.9  & 0.999 \\
\hline\\[-1em]
\textbf{Avg R-prec}   & 72.91 & 72.98 & 73.02 & 73.16 & 73.61  & 73.74 \\ \hline
\end{tabular}
\caption{Average page-level R-precision w.r.t momentum factor for our adaptive learning
}
\label{tab:beta}
}
\end{table}

Table~\ref{tab:temp} shows the impact of softmax temperature on validation R-precision for our adaptive learning. Table~\ref{tab:beta} shows the impact of momentum factor on validation R-precision for our adaptive learning.


\section{Passage-level Performance}\label{app:passage_level}
Table~\ref{tab:val_passage} shows the passage-level R-precision on KILT validation data. We also list the passage-level performance from \citet{maillard2021multi} for comparison.

\begin{table}[h!]
  \begin{adjustbox}{width=.48\textwidth,center}
    \begin{tabular}{lccccccc|c}
      \Xhline{2\arrayrulewidth}
      \rule{0pt}{1.1em}& Fact Check. & \multicolumn{2}{c}{Slot Filling}  & \multicolumn{3}{c}{Open Domain QA} & Dial. & \\
       \rule{0pt}{1.1em}{\bf Model} & {\bf FEV} &  {\bf T-REx} & {\bf zsRE} & {\bf NQ} & {\bf HoPo} & {\bf TQA} & {\bf WoW} & {\bf Avg} \\
           \hline\\[-1em]
        MT-DPR$^*$ & \underline{46.96} & 53.54 & 41.70 & 28.80 & 38.42 & 24.56 & 24.07 & 36.86 \\
        Task-specific DPR$^*$ & 43.92 & 58.54 & 78.81 & 28.13 & \underline{43.47} & 23.79 & 20.73 & 42.48 \\
        Task-specific (ours) & 44.89 & \underline{72.09} & {\bf 84.47} & {\bf 33.14} & 43.40 & {\bf 29.57} & \underline{27.64} & \underline{47.89} \\
   \hline
        TACO  & {\bf 60.76}  &  {\bf 72.57}  & \underline{82.80}  &  \underline{31.16}  & {\bf 46.72}  & \underline{28.32}  & {\bf 33.24}  & {\bf 50.80}  \\
      \Xhline{2\arrayrulewidth}
    \end{tabular}

  \end{adjustbox}
    \caption{Passage-level R-precision on KILT validation data.
    {\bf Bold} indicates the best  model and \underline{underline} indicates the second.  $*$  marks results from  \citet{maillard2021multi}.Only page-level R-precision is defined for AIDA. }
  \label{tab:val_passage}
\end{table}

\end{document}